\documentclass[letterpaper, 10pt, conference]{ieeeconf}  

\IEEEoverridecommandlockouts                              

\usepackage{xr}
\newcommand{\omitted}[1]{}
\usepackage{subfigure}
\usepackage{float}
\usepackage{graphicx}
\usepackage{xspace}
\usepackage{footmisc}

\usepackage[font=small,labelfont=bf]{caption}
\usepackage{siunitx}
\usepackage[absolute]{textpos}

\newcommand{\extended}[2]{{#1}\xspace} %

\title{\huge \bf \Aggressive Grasping  with a ``Soft'' Drone: \\ From Theory to Practice} 

\author{
		Joshua Fishman, Samuel Ubellacker, Nathan Hughes, Luca Carlone
		\thanks{J. Fishman, S. Ubellacker, N. Hughes, and L. Carlone are with the 
		Laboratory for Information and Decision Systems~(LIDS), Massachusetts Institute of Technology,
		        Cambridge, MA.} \thanks{ {\tt\footnotesize \{joshuaf,subella,na26933,lcarlone\}@mit.edu }} 
}

\usepackage{comment}
\usepackage{siunitx}
\usepackage{relsize}
\usepackage{ifthen}
\usepackage[colorinlistoftodos]{todonotes}

\usepackage[caption=false]{subfig}

\usepackage[vlined,ruled,linesnumbered]{algorithm2e}
\usepackage{graphics} 
\usepackage{rotating}
\usepackage{color}
\usepackage{enumerate}
\usepackage[T1]{fontenc}
\usepackage{psfrag}
\usepackage{epsfig} 
\usepackage{booktabs}
\usepackage{graphicx,url}
\usepackage{multirow}
\usepackage{array}
\usepackage{latexsym}
\usepackage{amsfonts}
\usepackage{amsmath}
\usepackage{amssymb}
\usepackage{xstring}
\usepackage[noend]{algorithmic}
\usepackage{multirow}
\usepackage{xcolor}
\usepackage{prettyref}
\usepackage{flexisym}
\usepackage{bigdelim}
\usepackage{breqn} 
\usepackage{listings}
\let\labelindent\relax
\usepackage{enumitem}
\usepackage{xspace}
\usepackage{bm}
\graphicspath{{./figures/}}
\usepackage{tikz}
\usetikzlibrary{matrix,calc}

\usepackage{mdwlist}

\let\itemize\undefined
\makecompactlist{itemize}{stditemize}

\newrefformat{prob}{Problem\,\ref{#1}}
\newrefformat{def}{Definition\,\ref{#1}}
\newrefformat{sec}{Section\,\ref{#1}}
\newrefformat{sub}{Section\,\ref{#1}}
\newrefformat{prop}{Proposition\,\ref{#1}}
\newrefformat{app}{Appendix\,\ref{#1}}
\newrefformat{alg}{Algorithm\,\ref{#1}}
\newrefformat{cor}{Corollary\,\ref{#1}}
\newrefformat{thm}{Theorem\,\ref{#1}}
\newrefformat{lem}{Lemma\,\ref{#1}}
\newrefformat{fig}{Fig.\,\ref{#1}}
\newrefformat{tab}{Table\,\ref{#1}}

\newcommand{\cf}{\emph{cf.}\xspace}

\newcommand{\bdmath}{\begin{dmath}}
\newcommand{\edmath}{\end{dmath}}
\newcommand{\beq}{\begin{equation}}
\newcommand{\eeq}{\end{equation}}
\newcommand{\bdm}{\begin{displaymath}}
\newcommand{\edm}{\end{displaymath}}
\newcommand{\bea}{\begin{eqnarray}}
\newcommand{\eea}{\end{eqnarray}}
\newcommand{\beal}{\beq \begin{array}{ll}}
\newcommand{\eeal}{\end{array} \eeq}
\newcommand{\beas}{\begin{eqnarray*}}
\newcommand{\eeas}{\end{eqnarray*}}
\newcommand{\ba}{\begin{array}}
\newcommand{\ea}{\end{array}}
\newcommand{\bit}{\begin{itemize}}
\newcommand{\eit}{\end{itemize}}
\newcommand{\ben}{\begin{enumerate}}
\newcommand{\een}{\end{enumerate}}

\newcommand{\calD}{{\cal D}}
\newcommand{\calE}{{\cal E}}

\newcommand{\calJ}{{\cal J}}

\newcommand{\setal}{~\emph{et~al.}\xspace}
\newcommand{\eg}{\emph{e.g.,}\xspace}
\newcommand{\ie}{\emph{i.e.,}\xspace}
\newcommand{\myParagraph}[1]{{\bf #1.}\xspace}

\newcommand{\M}[1]{{\bm #1}} 
\renewcommand{\boldsymbol}[1]{{\bm #1}}

\newcommand{\hide}[1]{}

\newcommand{\hiddenText}{{\color{gray} hidden text.}}
\newcommand{\hideWithText}[1]{\hiddenText}

\newcommand{\subject}{\text{ subject to }}
\DeclareMathOperator*{\argmax}{arg\,max}
\DeclareMathOperator*{\argmin}{arg\,min}

\newcommand{\tran}{^{\mathsf{T}}}

\newcommand{\diag}[1]{\mathrm{diag}\left(#1\right)}

\newcommand{\Real}[1]{ { {\mathbb R}^{#1} } }

\newcommand{\SOthree}{\ensuremath{\mathrm{SO}(3)}\xspace}

\newcommand{\MJ}{\M{J}}

\newcommand{\MR}{\M{R}}

\newcommand{\MX}{\M{X}}
\newcommand{\MY}{\M{Y}}

\newcommand{\vb}{\boldsymbol{b}}

\newcommand{\ve}{\boldsymbol{e}}
\newcommand{\vf}{\boldsymbol{f}}
\newcommand{\vg}{\boldsymbol{g}}

\newcommand{\vl}{\boldsymbol{l}}

\newcommand{\vo}{\boldsymbol{o}}
\newcommand{\vp}{\boldsymbol{p}}

\newcommand{\vy}{\boldsymbol{y}}

\newcommand{\vtheta}{\boldsymbol{\theta}}

\newcommand{\vtau}{\boldsymbol{\tau}}

\newcommand{\blue}[1]{{\color{blue}#1}}

\newcommand{\linkToPdf}[1]{\href{#1}{\blue{(pdf)}}}
\newcommand{\linkToPpt}[1]{\href{#1}{\blue{(ppt)}}}
\newcommand{\linkToCode}[1]{\href{#1}{\blue{(code)}}}
\newcommand{\linkToWeb}[1]{\href{#1}{\blue{(web)}}}
\newcommand{\linkToVideo}[1]{\href{#1}{\blue{(video)}}}
\newcommand{\linkToMedia}[1]{\href{#1}{\blue{(media)}}}
\newcommand{\award}[1]{\xspace} 

\graphicspath{{./figures/}}

\newcommand{\finalTime}{{t_f}}
\newcommand{\graspTime}{{t_g}}

\newcommand{\quadpos}{\vp}
\newcommand{\quadrot}{\MR}
\newcommand{\quadrvel}{\boldsymbol{\Omega}}
\newcommand{\propforces}{\vf}

\newcommand{\nodes}{\MY}
\newcommand{\node}{\vy}
\newcommand{\MXbar}{\bar{\MX}}
\newcommand{\MYbar}{\bar{\MY}}

\newcommand{\restlengths}{\vl}

\newcommand{\dt}{\text{d}t}

\newcommand{\rotcolx}{\vb_x}
\newcommand{\rotcoly}{\vb_y}
\newcommand{\rotcolz}{\vb_z}

\newcommand{\youngModulus}{\calE}
\newcommand{\poissonRatio}{\nu}

\newcommand{\softDrone}{{soft drone}\xspace}
\newcommand{\optional}[1]{}

\newcommand{\Aggressive}{Dynamic\xspace}
\newcommand{\aggressive}{dynamic\xspace}

\newcommand{\disturbance}{\vtheta}
\newcommand{\estDisturbance}{\hat{\disturbance}}

\newcommand{\feedforward}{feedforward\xspace}
\newcommand{\graspTarget}{target\xspace}

\PassOptionsToPackage{end}{algorithmic}

\renewcommand{\linkToPdf}[1]{\xspace}
\renewcommand{\linkToPpt}[1]{\xspace}
\renewcommand{\linkToCode}[1]{\xspace}
\renewcommand{\linkToWeb}[1]{\xspace}
\renewcommand{\linkToVideo}[1]{\xspace}
\renewcommand{\linkToMedia}[1]{\xspace}
\renewcommand{\award}[1]{\xspace}

\begin{document}

\maketitle

\begin{textblock}{10}(3,0.1)
\large \centering
\noindent Please cite as follows: J. Fishman, S. Ubellacker, N. Hughes and L. Carlone, 

\noindent "Aggressive Grasping  with a ``Soft'' Drone: From Theory to Practice"

\noindent  IEEE/RSJ International Conference on Intelligent Robots and Systems, 2021
\end{textblock}

\thispagestyle{empty}
\pagestyle{empty}


\begin{abstract}
Rigid grippers used in existing aerial manipulators require precise positioning to 
achieve successful grasps and transmit large contact forces that may destabilize the drone.  This limits the speed during grasping and prevents ``dynamic grasping'', 
where the drone attempts to grasp an object while moving. 
On the other hand, biological systems (\eg birds) rely on compliant and soft parts to dampen contact forces and {compensate for grasping inaccuracy,
enabling impressive feats.}

This paper presents the first prototype of a \emph{\softDrone}
--- a quadrotor where traditional (\ie~rigid) landing gears are replaced with a soft tendon-actuated gripper to enable aggressive grasping.
We provide three key contributions.
First, we describe our \softDrone prototype, including electro-mechanical design, software infrastructure, and fabrication.
Second, we review the set of algorithms we use for trajectory optimization and control of the drone and the soft gripper;
the algorithms combine state-of-the-art techniques for quadrotor control (\ie an {adaptive geometric controller}) with advanced soft robotics models (\ie a quasi-static finite element model).
Finally, we evaluate our \softDrone in physics simulations (using SOFA and Unity) and in real tests in a motion-capture room.
Our drone is able to dynamically grasp objects of unknown shape where baseline approaches fail.
Our physical prototype ensures consistent performance,
achieving 91.7\% successful grasps across 23 trials.
We showcase dynamic grasping results in the video attachment. 
\end{abstract}

Video Attachment: 
\blue{\footnotesize\url{https://youtu.be/mqbj8mEyCdk}}


\section{Introduction}
\label{sec:intro}

Quadrotors have been extensively investigated as platforms for navigation and inspection~\cite{Falanga18ral, Bodie19arxiv}, 
autonomous transportation and construction~\cite{Loianno18ral}, 
medical goods delivery~\cite{Thiels15amj},
agriculture and forestry~\cite{Ore13fsr},
among others~\cite{Khamseh18ras-aerialManipulationSurvey}.
State-of-the-art systems show impressive performance in agile navigation~\cite{Falanga18ral, Bodie19arxiv}, but are still relatively slow and overly-prudent when it comes to manipulation. 
Small quadrotors (often called \emph{micro aerial vehicles}~\cite{Loianno18ral}) have limited payload and can only carry relatively simple manipulators.
This intrinsically limits their capability to compensate for disturbances,
such as those caused by the quadrotor positioning errors during grasp execution. 
Moreover, rigid arms directly transfer external contact forces to the quadrotor, 
potentially causing instability 
during grasping~\cite{Khamseh18ras-aerialManipulationSurvey}. 
Many works avoid these issues by reducing speed during grasping~\cite{Dentler16cca-aerialManipulation,Rossi17ral}, 
which comes at the cost of inefficiency and increased operation times.  

\begin{figure}[t]
    \centering
    \includegraphics[width=\columnwidth]{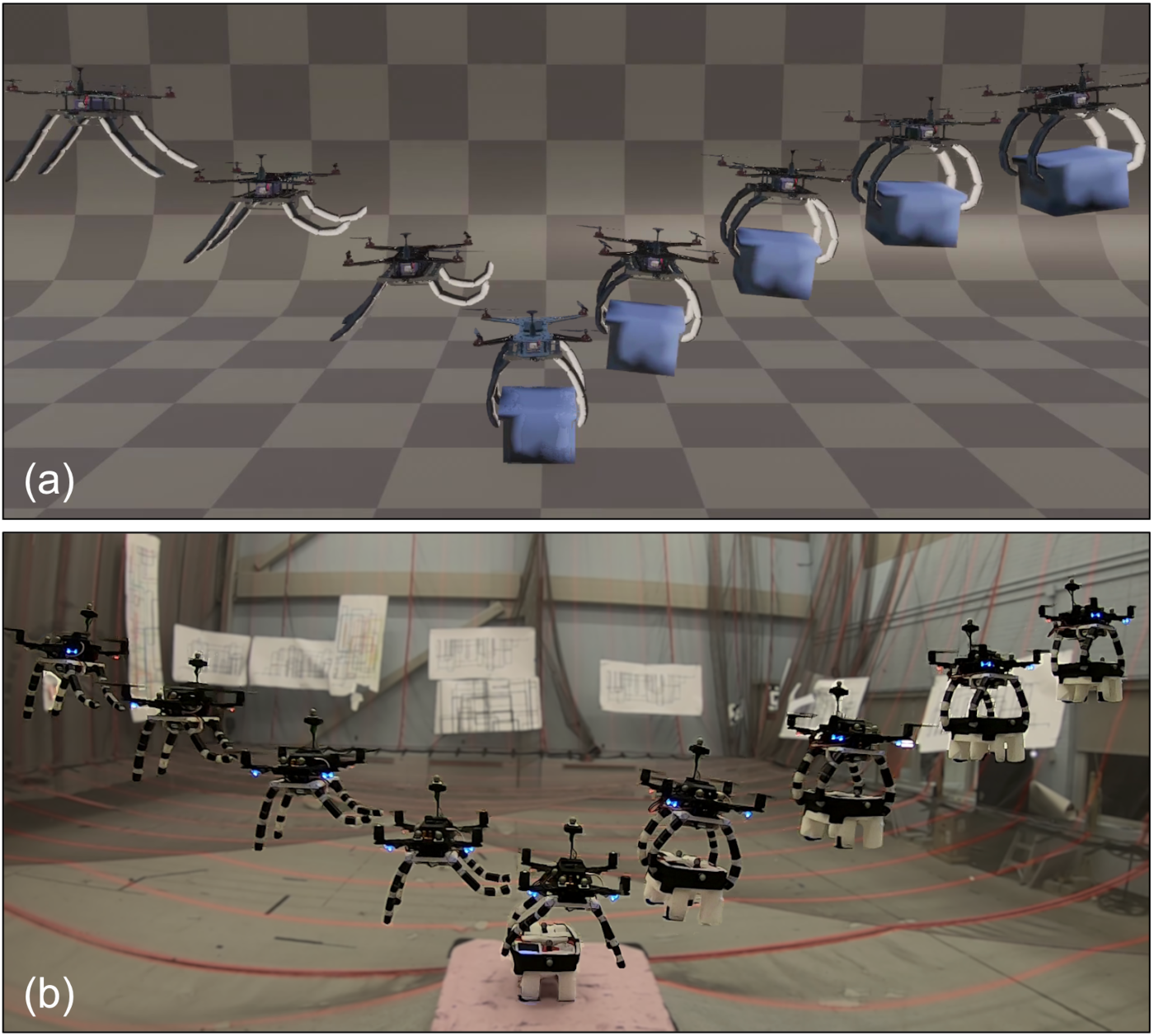}
    \caption{
    We develop the first prototype of a \emph{soft drone}, where 
  an actuated soft robot replaces the landing gear of a drone for dynamic grasping. 
    (a) Temporal sequence showing grasping at \SI{0.2}{m/s} in Unity-SOFA     physics simulator, 
    (b) Dynamic grasping in real tests. \label{fig:timelapse}}
\end{figure}

In contrast, biological systems (\eg birds) rely on compliance and softness to
enable the performance and robustness which so differentiate natural from artificial systems~\cite{Bern19rss}. 
Soft manipulators passively conform to the grasped object, enabling tolerance to imprecisions  and reducing the need for explicit grasp analysis; this is an example of \emph{morphological computation}, the exploitation of passive mechanical elements to supplement explicit control~\cite{Rus15nature}. 
However, the use of soft grippers in aerial manipulation poses new challenges: the system has infinite degrees of freedom and is no longer differentially flat.
Our previous work~\cite{Fishman21aeroconf-softDrone} provides an initial design to overcome these challenges
but only demonstrates the resulting system in simulation. In this paper, we take a step further
and present the first \emph{real prototype} of a \softDrone, demonstrating its potential in \aggressive grasping 
(Fig.~\ref{fig:timelapse}). 
We also make substantial changes to the design of the gripper and adopt an improved control scheme.

\myParagraph{Related Work}
Compliant aerial manipulation is a growing research area, with several contributions focusing on differentially flat
systems with limited degrees of freedom~\cite{yuksel16iros}.
Cable-slung loads~\cite{Foehn17rss,Sreenath13icra,Kotaru17acc} are a well-studied under-actuated payload, where
 the cable is either treated as massless or reduced to a finite number of links~\cite{Goodarzi14acc}. 
 Thomas\setal~\cite{Thomas14bioinspiration} carry out \aggressive grasping with a rigid under-actuated gripper (see also \cite{Pounds11icra,Backus14iros}), but focus on grasping a suspended object --- a setup that avoids unplanned contact forces. 
The AEROARMS project~\cite{Cabellero18iros,Ollero18ram} explores the use of a manipulator with a flexible link to minimize disturbance on the aerial platform but adds a passive joint to eliminate the impact of compliance on the drone dynamics.
Deng\setal~\cite{Deng20arxiv} implement and control a soft-bodied multicopter in simulation; their work relies on a gray-box neural network for system identification and does not consider manipulation.

Related work also investigates new quadrotor designs that adjust to different environments.  
Falanga\setal~\cite{Falanga18ral} develop a drone that folds its arms to fit into narrow gaps.
Ramon\setal~\cite{Ramon19iros} propose soft landing gear (similar in spirit to our design) but do not model or control the soft component beyond a binary open/close, and 
focus on perching rather than grasping.  
Mintchev\setal~\cite{Mintchev17ral} use insect-inspired structural compliance to minimize impact damage. 
Our paper connects aerial manipulation with the emerging discipline of \emph{soft robotics}~\cite{Rus15nature,Thuruthel18softrobotics}.
Soft robotics research has developed bio-inspired tendon-actuated soft grippers~\cite{King18humanoids,Manti15SoftRobotics,Hassan15embc}, 
and mathematical models to describe soft robots using finite element methods~\cite{Bern17iros,Bern19rss,Duriez13icra} and 
piecewise-constant-curvature approximations~\cite{Marchese14icra,Marchese15icra}.
These works have not been applied in the context of aerial grasping, 
where the soft gripper becomes a time-varying payload for a drone and impacts its dynamics.

\myParagraph{Contribution} 
This paper presents the first prototype of a \emph{\softDrone} (Fig.~\ref{fig:system-sim}), 
a quadrotor platform where traditional landing gear is replaced with a soft tendon-actuated gripper to enable \aggressive grasping. 
Our first contribution is to describe our \softDrone prototype (Section~\ref{sec:system_overview}), 
including electro-mechanical design, software infrastructure, and fabrication. 
Our system relies on a bio-inspired soft finger design, 
realized using lightweight tendon-actuated silicone rubber.

Our second contribution (Section~\ref{sec:aggressiveManipulation}) is to review the algorithms we use for trajectory optimization and control of the drone and the soft gripper. 
For the drone, we combine a minimum-snap trajectory optimizer~\cite{Mellinger11icra} with the 
\emph{adaptive geometric controller} of Goodarzi\setal~\cite{Goodarzi15jdsmc-adaptiveGeometricControl}. 
The adaptive controller is instrumental to enable grasping in real tests: 
it allows the drone to quickly adjust to changes in the quadrotor dynamics due to the swinging soft gripper and the mass of the object being grasped and
mitigates other state-dependent aerodynamic disturbances, such as the \emph{ground effect}~\cite{Bernard18aagnc-groundEffect}.
For the soft gripper, we use a \feedforward controller applied to the trajectory optimization scheme
 of~\cite{Fishman21aeroconf-softDrone}.

Our last contribution (Section~\ref{sec:experiments}) is an extensive evaluation of our system.
We evaluate our \softDrone in realistic simulations based on SOFA~\cite{Faure12sofaam} and Unity, and test our prototype in real 
grasping experiments.
In simulation, we compare the proposed platform against a rigid analogue and demonstrate the advantages of our choice of control algorithms and the advantage of softness.
In the real tests, we fly our \softDrone in a motion-capture room and show that it bears out its theoretical promise 
and is able to {dynamically grasp an object of unknown shape.}
We release \aggressive grasping 
videos in the supplementary material.

\section{Soft Drone Design and Prototyping}
\label{sec:system_overview}


\begin{figure}[tbp]
    \begin{minipage}{0.5\columnwidth}
        \begin{center}
            \includegraphics[trim={0cm 5mm 0cm 0cm},clip,width=0.99\columnwidth]{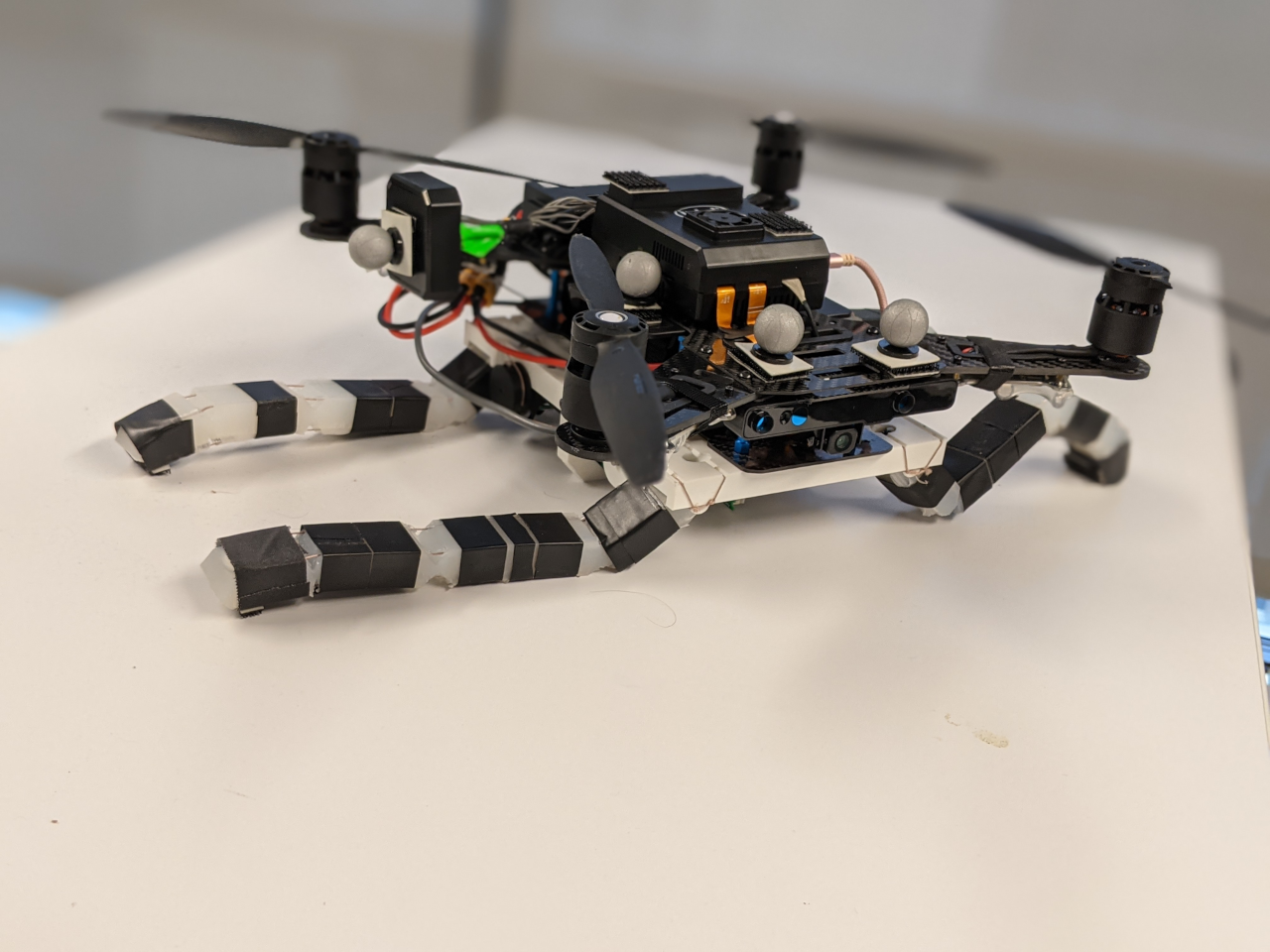} \\
        \end{center}
        \vspace{-8mm}
        \hspace{0.5em}\color{black}(a)
    \end{minipage}%
    \begin{minipage}{0.485\columnwidth}
        \begin{center}
            \includegraphics[trim={0cm 0cm 0cm 1.5cm},clip,width=0.99\columnwidth]{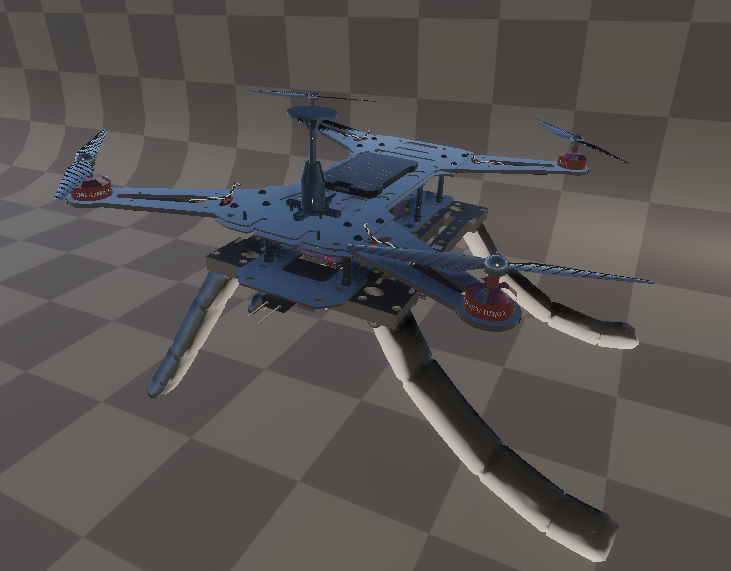} \\
        \end{center}
        \vspace{-8mm}
        \hspace{0.5em}\color{black}(b)
    \end{minipage}
    \vspace{2mm}
    \caption{Our \emph{\softDrone} is a standard quadrotor (the \emph{Intel Aero Ready To Fly} platform) retrofitted with a tendon-actuated soft gripper in place of the landing gear. (a) Our physical prototype. (b)~A~rendering of our {\softDrone} in the Unity-SOFA simulator.}\label{fig:system-sim} 
\end{figure}

Our \softDrone system consists of a standard quadrotor with the (rigid and heavy) landing gear replaced by a soft gripper (Fig.~\ref{fig:system-sim}).
The gripper comprises four silicone rubber fingers attached to the quadrotor base, and placed in a configuration similar to the rigid landing gear they replace; this configuration allows the fingers to  achieve force closure in an enveloping grasp and support a stable landing. The gripper is actuated by 16 tendons (similar to the
 design by {Hassan\setal~\cite{Hassan15embc}}) --- two on each side of each finger. However, there are only four unique tendon lengths: tendons on the same side of a finger are coupled to prevent finger twist while those on the same side of each pair of fingers oriented in the same direction are coupled to enforce planarity, simplifying planning{~\cite{Bern19rss}}. The quadrotor base uses four motors and propellers for actuation as usual.
In summary, the system uses 8 control variables: 4 motor speeds for the quadrotor, and 4 tendon lengths (later called the \emph{rest lengths}) for the gripper. 
These variables control the 6-dimensional quadrotor pose and the infinite-dimensional state describing the soft gripper configuration. 

\subsection{Quadrotor Base}

The quadrotor base of our soft drone 
is based on the \textit{Intel Aero Ready To Fly} quadrotor, which is  customizable and can be easily interfaced with
the Robot Operating System (ROS) and standard motion capture systems (\eg Optitrack or Vicon), which are used to supplement the quadrotor's state estimation and relay the location of the target object to grasp.
The quadrotor also features a relatively compact combination of an embedded real-time flight controller running the
PX4 Autopilot~\cite{px4} combined with a more capable companion computer (the \textit{Intel Aero Compute Board}) dedicated to running ROS. This setup allows us to naturally relegate low-level control (running at 250Hz) to the flight controller and execute polynomial trajectory optimization on the onboard computer at 50Hz~\cite{Fishman21aeroconf-softDrone}.

\subsection{Soft Gripper Design and Prototyping}
\label{sec:implementation_details}

\begin{figure}
    \centering
    \includegraphics[width=.38\textwidth]{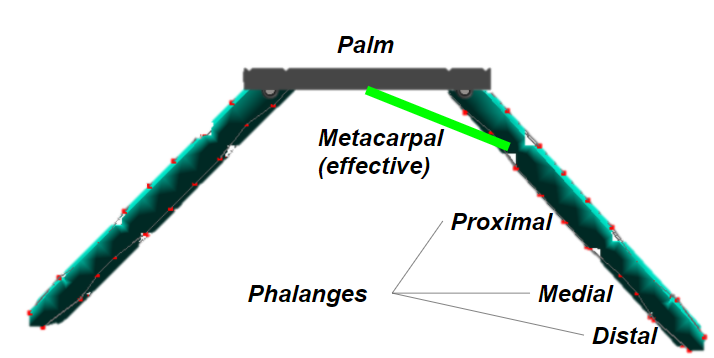}
    \caption{Our soft fingers (each $\SI{0.2}{m}$, green in the figure) are attached at a 45 degree angle to the $\SI{0.14}{m}$ base (dark gray). Tendons (black) are attached to mesh nodes (red). Each finger comprises metacarpal, proximal, medial, and distal segments in a 3.5:2.5:1.5:1 ratio, measuring the metacarpal from the middle of the base.
    \label{fig:finger_segments}}
\end{figure}

\myParagraph{Mechanical Design}
Our soft fingers are $\SI{0.2}{m}$ long and are attached to a $\SI{0.14}{m}$-wide base at a 45 degree angle (Fig.~\ref{fig:finger_segments}).
The fingers nominally deform continuously, but we add cutouts to the silicone to guide the deformations.
The joint placement is inspired by a human finger (a common methodology in robotic hand design \cite{Schlagenhauf18humanoids}): the three joints divide the base and finger into metacarpal, proximal, medial and distal segments (the latter three known collectively as ``phalanges''), which are sized in the same approximate 3.5:2.5:1.5:1 ratio as a human finger \cite{Buryanov10Morphology-fingerProportions}. Further, we restrict the nominal range of motion of the cutouts to 90, 90, and 30 degrees, similar to a human finger.

\myParagraph{Fabrication}
We mold our fingers using \textit{Smooth-On Dragon Skin 30} silicone rubber.
To route tendons through the fingers, low-friction nylon tubing is integrated into the mold and secured with electrical tape (Fig.~\ref{fig:system-sim}(a)). These components are attached to a 3D-printed base (Dremel Eco-ABS), which mounts directly to the quadrotor. Our tendons are 80lb-test braided fishing line (Power-Pro Super 8 Slick), running through the fingers and base to 3D-printed winches attached to our motors.

\myParagraph{Electronics}
The gripper is controlled by an \emph{Arduino Due} microcontroller, which receives tendon length commands from the \emph{Intel Aero Compute Board} on the quadrotor 
via USB and issues proportional commands to four motors 
via two \emph{TB6612 Motor Drivers} (with motor voltage via a 12V boost/buck regulator) on a custom shield. To control the outer tendons, we used two \emph{298:1 Pololu Micro Metal Gearmotors}; for the inner tendons, which are under higher loads, we use two 
\emph{250:1 Pololu 20D Metal Gearmotors}.


\label{sec:implementation_details}


\section{\Aggressive Grasping with a Soft Drone}
\label{sec:aggressiveManipulation}

Our \softDrone is tasked with grasping an object of unknown shape, given the coordinates of its centroid. 
We measure the  
state of the quadrotor (\eg from an onboard state estimator or a motion-capture system), 
but not that of the soft gripper. 
Let us call $\MX(t)$ the state of the quadrotor base (\ie a 3D pose and its derivatives) at time $t$,
and $\MY(t)$ the infinite-dimensional matrix describing the 3D position of every point of the soft gripper.
Call
$\propforces(t)$ 
the quadrotor propeller {thrust} forces at time $t$, and 
$\restlengths(t)$
the tendon lengths that actuate the fingers. 
Below, we omit the dependence on time $t$ when possible.
Our desired trajectory and control law can be formulated as an optimal control problem:
\begin{equation}
\begin{array}{rcl}
\hspace{-2mm} 
    (\MX^\star, \MY^\star, \propforces^\star, \restlengths^\star)  \label{eq:optControl}
     = \hspace{-9mm}& \hspace{-2mm}
     \displaystyle \argmin_{\MX, \MY, \propforces, \restlengths} &
    \int_{0}^{\finalTime}  \calJ(\MX, \MY, \propforces, \restlengths) \dt
    \\
    & \hspace{-4mm}  \subject & 
        \calD(\MX, \MY, \propforces, \restlengths) = 0 \\
    && \MX(0) = \MXbar_0, \quad \MY(0) = \MYbar_0 \\ 
    && \MX(\finalTime) = \MXbar_\finalTime, \; \MY(\finalTime) = \MYbar_\finalTime \\
     && \MY(\graspTime) = \MYbar_\graspTime
\end{array}
\end{equation}
where $\calJ(\MX, \MY, \propforces, \restlengths)$ is the cost functional that, for instance, penalizes 
control usage or encourages smooth state changes,
 the constraint $\calD(\MX, \MY, \propforces, \restlengths)=0$ ensures that the solution satisfies 
 the platform dynamics,
$(\MXbar_0,\MYbar_0)$ is the given initial state of the soft aerial manipulator at the initial time $t=0$, 
$(\MXbar_\finalTime,\MYbar_\finalTime)$ is the desired state at the final time $t_f$ (say, the end of the execution), 
and $\MYbar_\graspTime$ is the desired state of the soft gripper at the time of grasp $\graspTime \in [0,\finalTime]$. 

Obtaining a control policy by directly solving~\eqref{eq:optControl} is impractical even without the soft
gripper~\cite{Bry15ijrr-aggressiveFlight}. The situation is exacerbated in our grasping problem, since 
(i) our system is not differentially flat due to the soft gripper, 
(ii) the grasp induces unmodeled contact forces,
and
 (iii) we cannot neglect aerodynamic effects when operating near surfaces.
 Therefore, 
 we first compute a nominal state trajectory via 
\emph{trajectory optimization}~\cite{Bry15ijrr-aggressiveFlight} and then design a controller to track such a trajectory while rejecting unmodeled disturbances. 
In particular, we use a minimum-snap polynomial optimization to find the optimal quadrotor trajectory, which informs a \emph{finite element} model to compute desired tendon lengths at key instants~\cite{Fishman21aeroconf-softDrone}. Then, we apply an \emph{adaptive controller} to track the quadrotor trajectory, while the tendon lengths are interpolated in open loop. More details are provided below.

\subsection{Soft Drone Control and Trajectory Optimization} 
\label{sec:controlAndPlanning}

We use the decoupled trajectory optimization approach we proposed in~\cite{Fishman21aeroconf-softDrone} but adopt a more advanced control scheme.
Intuitively, we first solve a trajectory optimization problem for the quadrotor to ensure it reaches a feasible grasp pose, and then we solve a trajectory optimization problem for the soft gripper given the quadrotor trajectory. 
\optional{We use a simple feedforward strategy for the control of the soft gripper, and adopt an adaptive controller for the quadrotor.}

\myParagraph{Quadrotor Trajectory Optimization and Control} 
We use a polynomial trajectory optimization approach~\cite{Bry15ijrr-aggressiveFlight,Mellinger11icra} to compute a minimum-snap trajectory that
starts at the initial drone position,
ends at a desired final position at the terminal time $t_f$,
and passes through an intermediate waypoint a finger-length above the target object at the time of grasp $t_g$.
We denote the nominal drone trajectory resulting from the minimum-snap trajectory optimization as $\MX^\star$.
Our previous work~\cite{Fishman21aeroconf-softDrone} used a simple geometric controller to track the nominal trajectory $\MX^\star$
in simulation, and
the presence of the soft gripper is treated as an unknown disturbance. 
The experiments in this paper show serious challenges to that approach in real \aggressive grasping scenarios. 
First, after grasping, the mass of the grasped object becomes a constant disturbance which causes a 
constant offset in the tracking error.
More importantly, real problems entail state-dependent aerodynamic disturbances that 
are not modeled in common simulators. 
The \emph{ground effect}, which is the extra lift incurred by 
a quadrotor moving close to the ground, becomes visible when our drone approaches the surface the target object lies on; if uncompensated it induces a positive offset in the vertical position of the drone, pushing it too far above the target object to grasp.
To address these issues, we adopt the \emph{adaptive} geometric controller of Goodarzi\setal~\cite{Goodarzi15jdsmc-adaptiveGeometricControl}, which we review below.

The quadrotor dynamics are modeled as: 
\begin{equation}
\left\{
\begin{array}{ll}
    m \Ddot{\quadpos} = m \vg + f \rotcolz + \disturbance_f \\
    \dot{\quadrot} = \quadrot \hat{\quadrvel} \\
    \MJ \dot{\quadrvel} = - \quadrvel \times \MJ \quadrvel + \vtau + \disturbance_\tau\\
\end{array}
\label{eq:quadrotor_dynamics}
\right.
\end{equation}
where 
the quadrotor state 
 $\MX \triangleq \{ \quadpos, \quadrot, \dot{\quadpos}, \quadrvel \}$ includes the 
quadrotor position $\quadpos \in \Real{3}$, 
 rotation $\quadrot\in \SOthree$,
 linear velocity $\dot{\quadpos} \in \Real{3}$, 
 and angular velocity $\quadrvel \in \Real{3}$, 
 and we denoted 
  the columns of $\quadrot$ as $\quadrot = [\rotcolx \; \rotcoly \; \rotcolz]$; 
  in~\eqref{eq:quadrotor_dynamics}, $m$ is the total mass of the platform and gripper, 
$\vg$ is the gravity vector, 
$\MJ$ is the moment of inertia, 
$f$ is the scalar thrust force (applied at the quadrotor center of mass and along the local vertical direction $\rotcolz$), 
and $\disturbance_f, \disturbance_\tau$ are (possibly state-dependent) disturbances induced by our soft gripper.

According to~\cite{Goodarzi15jdsmc-adaptiveGeometricControl}, we compute the control actions (\ie the thrust magnitude and the moment vector) as:
\bea
\label{eq:adaptiveGeoController}
 f &=& -\rotcolz\tran ( \overbrace{k_p \ve_p + k_v \ve_v+ 
                        m \vg - m \Ddot{\quadpos}_d}^{geometric} + 
                        \overbrace{\estDisturbance_f}^{adaptive})  \\
    \vtau &=& -k_r \ve_r - k_{\Omega} \ve_{\Omega} + \quadrvel \times \MJ \quadrvel \nonumber \\
          & & \underbrace{- \MJ( \hat{\quadrvel} \quadrot\tran \quadrot_d \quadrvel_d - \quadrot\tran \quadrot_d \dot{\quadrvel}_d     )}_{geometric} - \underbrace{\estDisturbance_\tau}_{adaptive}
\eea
where $\quadpos_d, \quadrot_d, \quadrvel_d$ are the desired position, attitude, and angular velocity, respectively,
$\ve_p,\ve_v,\ve_r,\ve_{\Omega}$ are the position, velocity, rotation, and angular velocity errors, 
$\estDisturbance_f, \estDisturbance_\tau$ are estimated disturbances on the translation and rotation dynamics,
and $k_p,k_v,k_r,k_{\Omega}$ are user-specified control gains.
In~\eqref{eq:adaptiveGeoController}, the geometric terms describe a proportional-derivative action 
(plus extra terms compensating for the desired state) and are the same used in a standard geometric controller~\cite{Fishman21aeroconf-softDrone,Lee10cdc-geometricControl}.
However, the control law~\eqref{eq:adaptiveGeoController} also includes adaptive terms that aim 
at learning the external disturbances $\disturbance_f,\disturbance_\tau$.

Assuming bounded disturbance $\|\disturbance_f\| \leq\!\beta$, $\|\disturbance_\tau \| \leq\!\beta$ (where $\beta$ is a given constant), the adaptive terms are adjusted online using the following law:
\bea
\label{eq:adaptiveLaw}
\frac{d\estDisturbance_f}{dt}\!=\! \Pi\left( \gamma_f (\ve_v \!+\! k_{af} \ve_p) \right), \;\;\;\; 
\frac{d\estDisturbance_\tau}{dt} \!=\! \gamma_\tau (\ve_{\Omega} \!+\! k_{a\tau} \ve_r) 
\eea
where $\gamma_f, \gamma_\tau, k_{af}, k_{a\tau}$ are user-specified gains   
and $\Pi$ is a suitable projection function.
Eq.~\eqref{eq:adaptiveLaw} estimates the time-varying disturbances to explain residual translation and rotation errors, but projects these estimates to keep them bounded in 
the ball of radius $\beta$ and avoid their divergence. 

Goodarzi\setal~\cite{Goodarzi15jdsmc-adaptiveGeometricControl} prove that this adaptive geometric controller 
ensures stability and its tracking errors asymptotically go to zero for suitable control gains.
In our setup, it enables accurate tracking despite the unmodeled mass of the grasped 
\mbox{object and 
 aerodynamic effects (\eg drag, ground effect).}

\myParagraph{Soft Gripper Optimization and Control} 
\label{sec:grasp_optimization_details}
We compute a nominal trajectory for the gripper given the nominal quadrotor base trajectory, by solving for the desired configuration of the gripper as well as the optimal tendon lengths over time. Details of our approach can be found in \cite{Fishman21aeroconf-softDrone}. 
We divide the grasp into two phases: ``approach'' (opening the gripper as the quadrotor approaches the target location, so as to allow the fingertips to surround the grasped object) and ``grasp'' (contracting the fingertips to achieve an enveloping grasp).

In particular, we compute the optimal configuration at the approach time $t_a$ (the moment when the leading fingertips are first above the target centroid, which can be determined from the nominal base trajectory  $\MX^\star$)
by solving the following optimization problem:
\beq
\label{eq:approach}
\MY^\star_a = \argmax_{\MY} \textstyle\sum_{i=1}^3 \| (\node_{tip_{i}} - \vo) \times  (\node_{tip_{i+1}} - \vo) \|_2^2
\eeq
where $\vo$ is the position of the grasp target and $\node_{tip_{i}} \in \MY$ are the fingertip positions of the gripper.
The cross product encourages the gripper to open the fingers and results 
in an asymmetric configuration (Fig.~\ref{fig:grasp_real}) where the fingers nearest to the 
target are lifted, while those farther  envelop the target.

Similarly, we compute the optimal configuration at the given time of grasp $t_g$ 
 by solving the following optimization:
 \beq
\label{eq:grasp}
\MY^\star_g = \argmin_{\MY} \textstyle\sum_{i=1}^4 \| \node_{tip_{i}} - \vo \|_2^2
\eeq
which simply tries to move the fingertips as close as possible to the object centroid. 


We remark that the optimization problems~\eqref{eq:approach} and~\eqref{eq:grasp} are nontrivial to solve
since they involve the continuously-deformable soft state $\MY$. Our approach follows Bern\setal\cite{Bern17iros}, approximating the infinite-dimensional soft gripper configuration as a set of discrete nodes $\nodes$ arranged in a \emph{tetrahedral mesh}, as in \emph{finite element methods} (FEM). 
Tendons are approximated as one-sided springs with (un-stretched) rest lengths $\restlengths$ which we control directly. A set of pins (also modeled as linear springs) fix the mesh nodes to the quadrotor base, and mass is approximated as concentrated in the nodes. We assume that the soft gripper is \emph{quasi-static}, \ie there is an instantaneous relation between rest lengths $\restlengths$ and gripper configuration $\MY$.

We iteratively solve for the lengths $\restlengths^\star$ yielding an optimal configuration $\MY^\star$. Starting from an initial guess $\bar{\restlengths}$ and $\bar{\nodes}$ (the rest configuration in our tests), we analytically compute a descent direction from the Jacobian $\frac{d\nodes}{d\restlengths}$ (details in \cite{Fishman21aeroconf-softDrone,Bern17iros})
and search along it\footnote{In practice we use parameters $\alpha=0.5$ and $\beta=0.1$ in computing the \textit{Armijo-Goldstein rule} \cite{Nocedal99} in backtracking line search along the gradient, as well as enforcing a step size norm between $1$ and $\SI{10}{cm}$. Further, we choose a learning rate of 2.5 when computing the approach lengths and 0.25 when computing the grasp lengths.}
for a suitable step  $\Delta \restlengths$.
Next, we find the new \emph{quasi-static} configuration $\nodes$ associated with the new tendon rest lengths $\restlengths = \bar{\restlengths} + \Delta \restlengths$
by minimizing the total system energy using Newton's method.\footnote{We use the \textit{Newton-CG} implementation in \textit{scipy.optimize.minimize}, with $\bar{\nodes} + \frac{d\nodes}{d\restlengths} \Delta \restlengths$ as an initial guess.}
The process is iterated till convergence, yielding a configuration $\nodes$ and tendon lengths $\restlengths$ which are approximately optimal.



Finally, between the instants for which we explicitly calculate tendon lengths (\ie $t_a$ and $t_g$), we perform a simple linear interpolation between the known lengths we compute above.
This is shown to minimize the $\ell_\infty$ norm of the time-derivative of the tendon length in~\cite{Fishman21aeroconf-softDrone}, hence penalizing rapid changes in tendon lengths (which would result in undesirably large forces on the tendon attachment points).

In summary, we solve explicitly for tendon lengths at times $t_a$ (right before grasping) and $t_g$ (time of grasp) which yield configurations $\MY^\star_a$ and $\MY^\star_g$.  We choose $\MY^\star_a$ to maximize the area between fingertips and the target given the quadrotor pose at that instant (eq.~\eqref{eq:approach}), thus maximizing the chance of an enveloping grasp. We choose $\MY^\star_g$ to bring the fingertips as close as possible to the target centroid (eq.~\eqref{eq:grasp}). Until time $t_a$ we linearly interpolate from initial to ``approach'' lengths; from $t_a$ to $t_g$ we interpolate from ``approach'' to ``grasp'' lengths; after $t_g$, tendon lengths are held constant; throughout, tendon lengths are applied in open loop.

\section{Experiments}
\label{sec:experiments}

We evaluate the effectiveness of our soft drone in photo-realistic simulations and in real tests. 
Our experiments show that
(i) in simulation, the adaptive controller converges regardless of the soft payload 
and ensures smaller errors compared to a standard geometric controller,
(ii) our system reliably achieves aggressive grasps (at up to 2\SI{}{m/s}) of objects of unknown shape while a rigid manipulator fails,
(iii) in real tests, our soft drone ensures successful grasps in 21 out of 23 consecutive tests, 
despite unmodeled aerodynamic effects.

\subsection{Photo-realistic Physics Simulations}
\label{sec:experimental_setup}

\myParagraph{Unity-SOFA Simulation Setup}
We simulate our soft drone in Unity, a popular 3D game engine. The underlying physics simulation uses the Unity plugin~\cite{SofaUnity} for SOFA~\cite{Faure12sofaam}, a popular open-source soft dynamics simulator with dedicated plugins for tendon-actuated soft manipulators~\cite{Duriez13icra}.

Unless otherwise noted, we use a simulation and control timestep of 0.01\SI{}{s}.
The rigid frame of the manipulator is modeled
after the frame of the \emph{Intel Ready to Fly quadrotor} (size: $0.25\times0.25\times0.04$\SI{}{m}),
while the four fingers are modeled as described in Section~\ref{sec:system_overview} (each with
size: $0.2 \times 0.02 \times 0.025\,\SI{}{m}$).
We choose quadrotor mass $m = \SI{1.7}{kg}$ and inertia $\MJ = \diag{[0.08, 0.08, 0.14]}
\SI{}{kg \cdot m^2}$ and assume aerodynamic drag on the quadrotor center of mass with body drag coefficient $0.3$.
For the soft components, we attempt to replicate the material properties of \textit{Smooth-On Dragon Skin 30} in simulation: we use
Young's modulus $\youngModulus= \SI{1}{MPa}$,
Poisson's ratio $\poissonRatio=0.25$,
Lam\'e parameters
$\mu = \frac{\youngModulus}{ 2 (1 + \poissonRatio)} \!=\! 400000 \SI{}{N/m^2}$, 
$\kappa=\frac{\poissonRatio \youngModulus}{(1 + \poissonRatio) (1 - 2\poissonRatio)} \!=\! 333333 \SI{}{N/m^2}$~\cite{Sifakis12siggraph},
and density $\rho = 1000 \SI{}{kg/m^3}$. 
The controller gains are set to  $k_p \!=\! 10, k_v \!=\! 10, k_r \!=\! 16, k_{\Omega} \!=\! 2.5, \gamma_f \!=\! 15, \gamma_\tau \!=\! 15, k_{af} \!=\! 2, k_{a\tau}=2$.
 Our simulated grasp target weighs 100 grams, approximately the same as our intended real target.

\begin{figure}[tbp]
    \centering \vspace{-3mm}
    \includegraphics[width=\columnwidth]{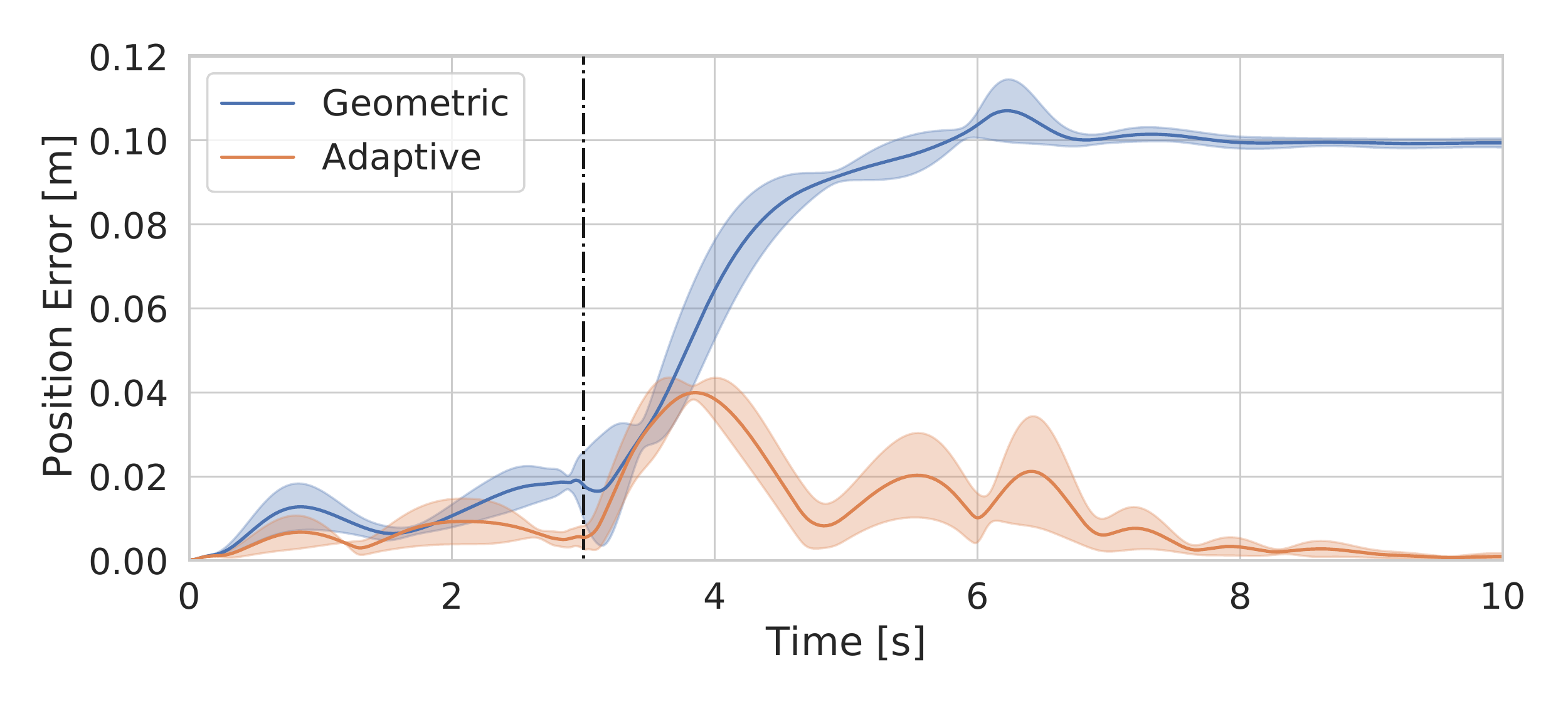} 
    \caption{Position tracking errors for the adaptive and standard geometric controller during grasping. Statistics are computed over 10 runs with different grasp velocities ranging from \SI{0.1}{m/s} to \SI{1}{m/s} in \SI{0.1}{m/s} increments. The dashed black vertical line denotes the time of grasp ($t_g = $\SI{3}{s})}\label{fig:controller_error}
\end{figure}

\myParagraph{Ablation: Quadrotor Controller}
Our first ablation study compares the adaptive controller with a standard geometric controller (Fig.~\ref{fig:controller_error}). While the original adaptive controller paper already reports a thorough evaluation for general use~\cite{Goodarzi15jdsmc-adaptiveGeometricControl}, we specifically compare the two controllers while grasping: we plan grasp trajectories across a range of velocities (\SI{0.1}{m/s} to \SI{1.0}{m/s}) and measure tracking errors. The resulting statistics for the position error are given in Fig.~\ref{fig:controller_error} (shaded area shows the 1-sigma deviation of the errors).
The adaptive controller has lower error throughout the trajectory, but much better performance after grasping the \graspTarget; in particular ---after grasping--- the tracking error of the adaptive controller converges to zero (as predicted by the convergence results in~\cite{Goodarzi15jdsmc-adaptiveGeometricControl}), while the geometric controller exhibits a significant 
steady state error (around \SI{0.1}{m}) caused by the presence of the unmodeled mass of the grasped object.

\myParagraph{Ablation: Rigid vs. Soft for Aggressive Grasping}
We compare our soft drone  against a 2-DOF rigid gripper of identical dimensions with a pre-programmed open-close action executed in open loop. To prevent the simulated gripper from clipping through the target, in these experiments we decrease the simulation timestep to $\SI{5}{ms}$ but maintain the 100\SI{}{hz} control rate.

\begin{figure}[tbp]
    \centering\vspace{-3mm}
    \includegraphics[width=\columnwidth]{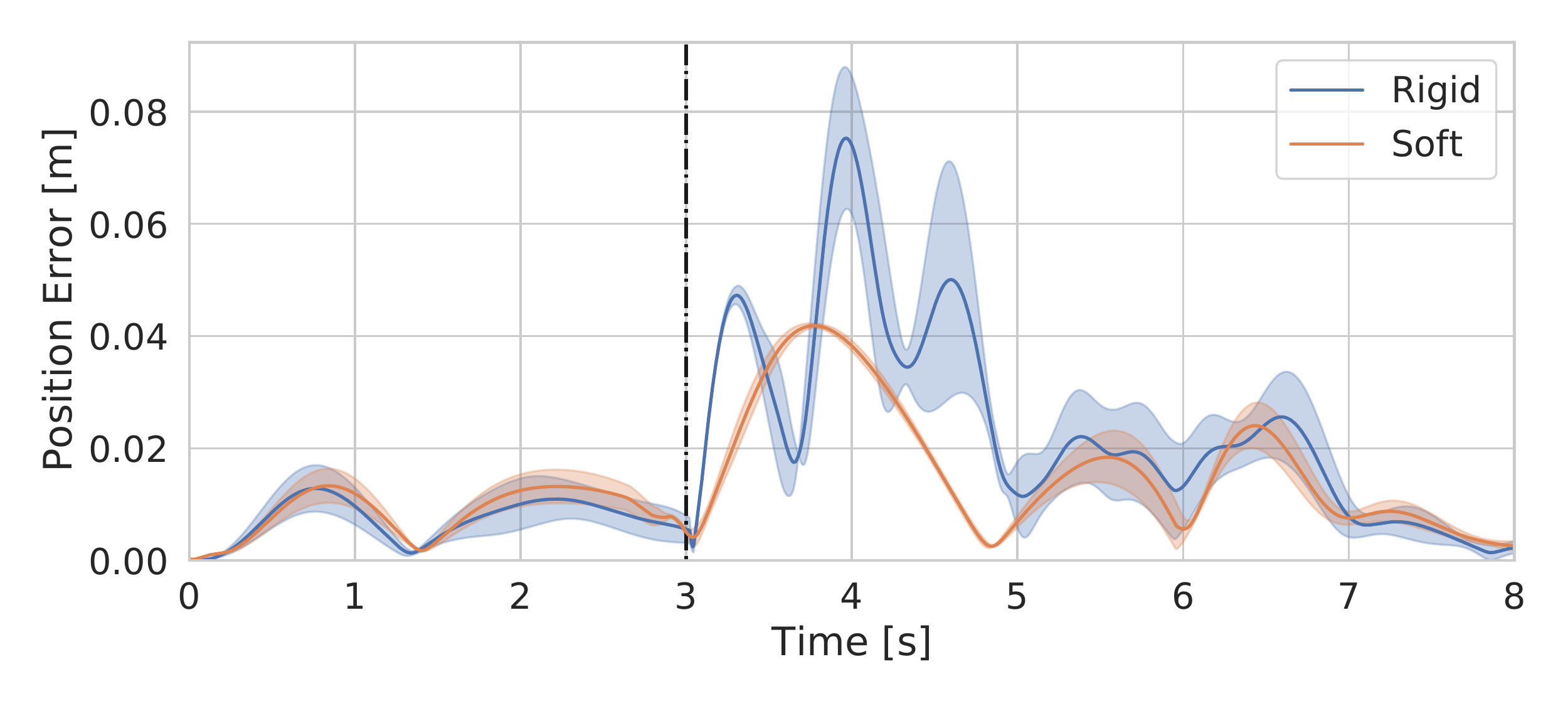}
    \caption{Position tracking error during successful grasp trajectories for both our soft drone and a rigid gripper design. Desired forward velocity at grasp is \SI{0.5}{m/s}. The dashed black vertical line denotes time of grasp ($t_g = $\SI{3}{s})}\label{fig:grasp_disturbances}
\end{figure}

Fig.~\ref{fig:grasp_disturbances} reports tracking errors during successful grasp trajectories for the soft and rigid gripper.
The soft gripper leads to 50\% smaller peak errors compared to its rigid counterpart: 
the soft gripper passively dampens the contact forces during grasping leading to smaller 
trajectory perturbations.

\begin{figure}[tbp]
    \centering
    \includegraphics[width=\columnwidth]{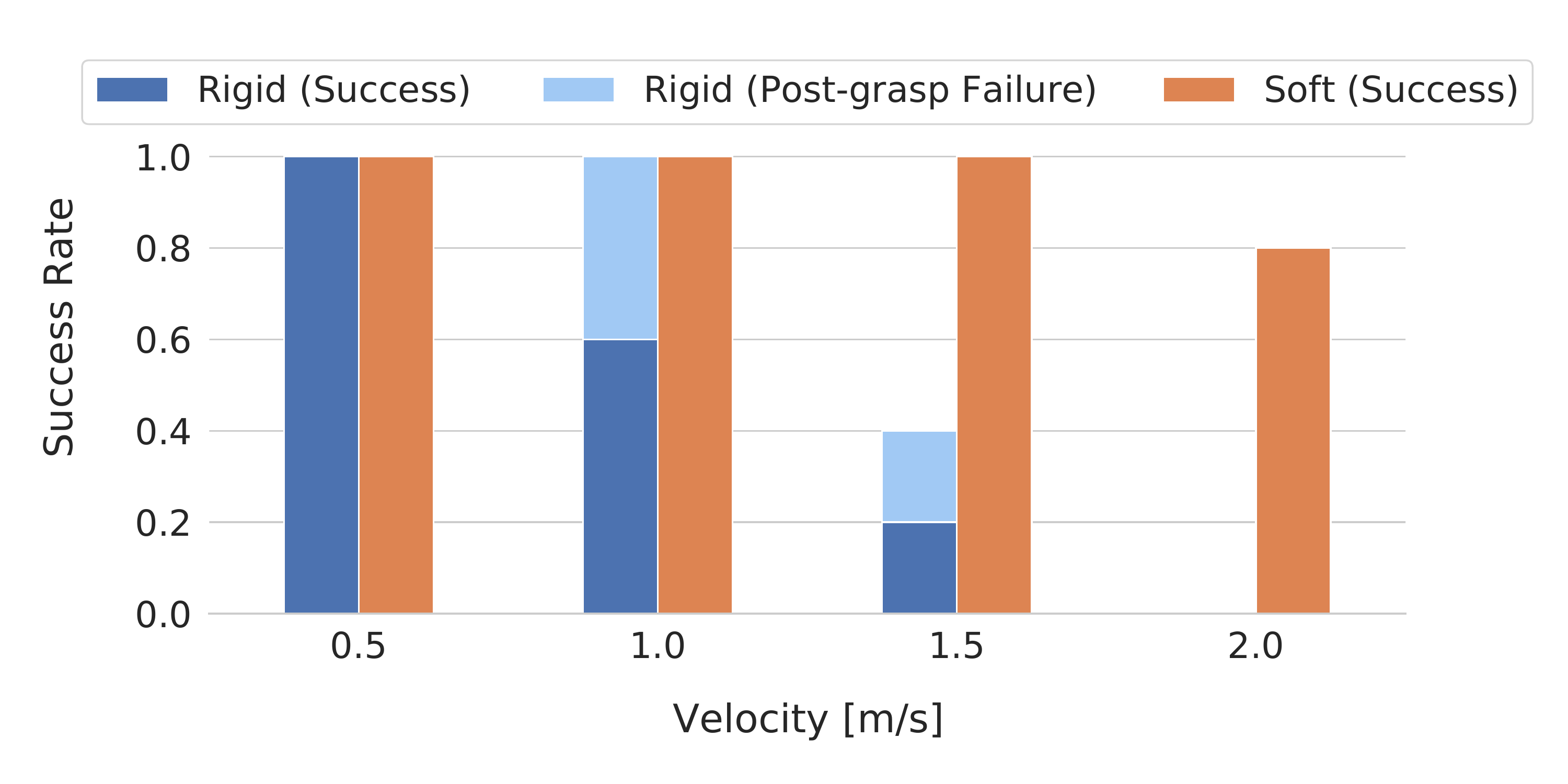}
    \caption{Grasp success rate for both our approach and a rigid gripper design across five trials for four different grasp velocities. ``Rigid (success)'' denotes that the rigid gripper achieved  successful grasping and reached the desired final position, while ``Rigid (Post-grasp Failure)'' denotes that the system successfully grasped the \graspTarget, but then the trajectory tracking controller diverged.}\label{fig:grasp_success}
\end{figure}

Fig.~\ref{fig:grasp_success} evaluates the success rate of the soft drone compared to a rigid gripper.
We simulated 5 trials of grasping at 4 different desired grasp velocities (\SI{0.5}{m/s}, \SI{1}{m/s}, \SI{1.5}{m/s}, \SI{2}{m/s}) and recorded the fraction of tests in which the drone remains stable and grasps the object. Our soft drone outperforms its rigid analogue at high velocities, 
and in particular the success of the rigid gripper drops to 40\% (against 100\% of the soft) at \SI{1.5}{m/s}, 
and all ``rigid'' tests fail at \SI{2}{m/s}.  
In general, the rigid gripper fails because it either makes an unexpected contact with the target or because the very large contact forces at higher velocities destabilize the controller; the magnitude of the contact forces is further exacerbated by the limitations of the 
SOFA simulator as discussed below. 
The soft gripper, on the other hand, 
absorbs large contact forces and consistently achieves grasping. 

\myParagraph{Limitations of Soft Simulation}
We observed several limitations in our physics simulation setup. As mentioned above, clipping and penetration between objects are common in SOFA at high velocities. 

Moreover, rigid-to-rigid contacts exhibit extremely large instantaneous forces despite our attempts to tune the simulator; as a consequence, the high-velocity grasps (in particular when using the rigid gripper)
vary in stability with the simulation timestep. This is the reason why we reported 
the post-grasp failures in Fig.~\ref{fig:grasp_success} as a separate category. 
Additionally, our simulator does not model complex \mbox{aerodynamic effects 
such as the ground effect.}

\begin{figure}[tbp]
    \centering
    \includegraphics[width=\columnwidth, trim={3.2cm 0cm 3.2cm 0cm}, clip,]{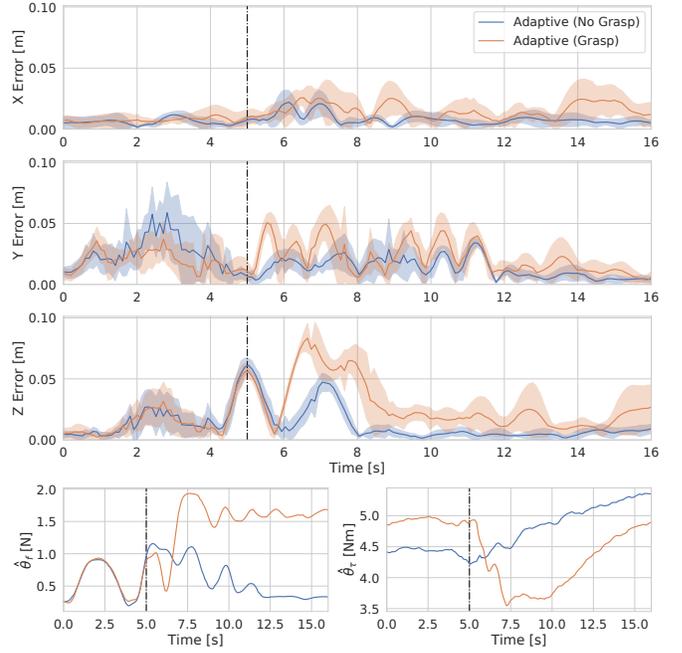}
    \caption{Position tracking error along the X, Y, Z axis during a grasp trajectory for the implemented adaptive controller when both attempting to grasp an object (``Grasp'') and simply tracking the grasp trajectory (``No Grasp'') across 5 trials. The bottom subplot displays the norm of the adaptive terms $\estDisturbance_f$ and $\estDisturbance_\tau$.\footref{fnlabel} Desired forward velocity at grasp is \SI{0.2}{m/s}. The dashed black vertical line denotes time of grasp ($t_g = $\SI{5}{s})}\label{fig:controller_error_real}
\end{figure}

\subsection{Real evaluation}

\myParagraph{Real testing setup}
We evaluate our real system using an Optitrack motion capture system for state estimation, communicating with the \textit{Intel Aero} drone via ROS. Additionally, we implemented a custom version of PX4 that replaced the original position and attitude controllers with the adaptive controller formulation of~\cite{Goodarzi15jdsmc-adaptiveGeometricControl}. The controller gains for all experiments were set to $k_p = 7.5, k_v = 6.0, k_r = 80.0, k_{\Omega} = 8.0, \gamma_f=10, \gamma_\tau=10, k_{af}=2, k_{a\tau}=2$.\footnote{Note that due to how PX4 manages the mapping between desired wrench and motor speeds, we were required to scale components of the wrench by an arbitrary factor. This resulted in the order of magnitude larger attitude gains between the real system and the simulated one.\label{fnlabel}} 
{We omit the derivatives of the velocity errors and the second derivative of the adaptive term 
$\estDisturbance_f$ from the computation of the angular velocity ${\quadrvel}_d$ and acceleration $\dot{\quadrvel}_d$ since we do not directly estimate jerk and acceleration (and the raw measurements from the IMU are too noisy).} 
We manually calibrate the static translational offset between the drone body frame as estimated by Optitrack and the drone's center of mass to be $\{0.02, 0.03, -0.03\}$. 
Initial tendon rest lengths are calibrated to be $\{190, 190, 208, 208\}\,\SI{}{mm}$.
In th{}e tests, we precomputed tendon lengths offline as in Section~\ref{sec:aggressiveManipulation} (which requires $\approx \SI{1}{s}$ \cite{Fishman21aeroconf-softDrone}); all other computation takes place onboard.
Our real grasping experiments use a foam target weighing $\SI{106}{g}$. Our combined quadrotor and gripper weighs $\SI{1.9}{kg}$, while the maximum load for our quadrotor is $\SI{2.0}{kg}$. 
We observe good grasping performance despite the fact that we operate near the maximum payload.

\myParagraph{Impact of Unmodeled Aerodynamic Effects}
As in simulation, we first evaluate the tracking errors for our system comparing the case 
where it grasps an object against the case in which it executes the same trajectory without any object to grasp. Fig.~\ref{fig:controller_error_real} reports statistics over 5 runs.
 In contrast to simulation, our real tests manifest an additional source of error: 
 the ``ground effect'' ---created by the platform our target rests on--- pushes the quadrotor upward during grasp. The influence that the ground effect is clearly visible from Fig.~\ref{fig:controller_error_real}: while the position errors in the horizontal plane (X,Y) are mostly low before grasping ($<\SI{5}{s}$), the vertical error (Z) rapidly increases right before grasping ($t_g = \SI{5}{s}$), when the drone approaches the surface the object is laid on.
  Note that the ground effect 
 impacts both case (\ie grasp or no grasp), while the errors induced by grasping become 
 visible after $t_g = \SI{5}{s}$. 
 The bottom plot in Fig.~\ref{fig:controller_error_real} reports the 
norm of the adaptive terms $\estDisturbance_f$ and $\estDisturbance_\tau$ and shows that
(i) the term $\estDisturbance_f$ increases before $t_g=\SI{5}{s}$ to counteract the ground effect, 
and (ii) the steady state value of $\estDisturbance_f$ is larger after grasping since it compensates
 for the mass of the grasped object (similarly $\estDisturbance_\tau$ compensates for unmodeled torques from the object).

 \begin{figure*}[tbp]
    \centering
    \includegraphics[trim={0cm 9.0cm 0cm 0.4cm}, clip, width=\textwidth]{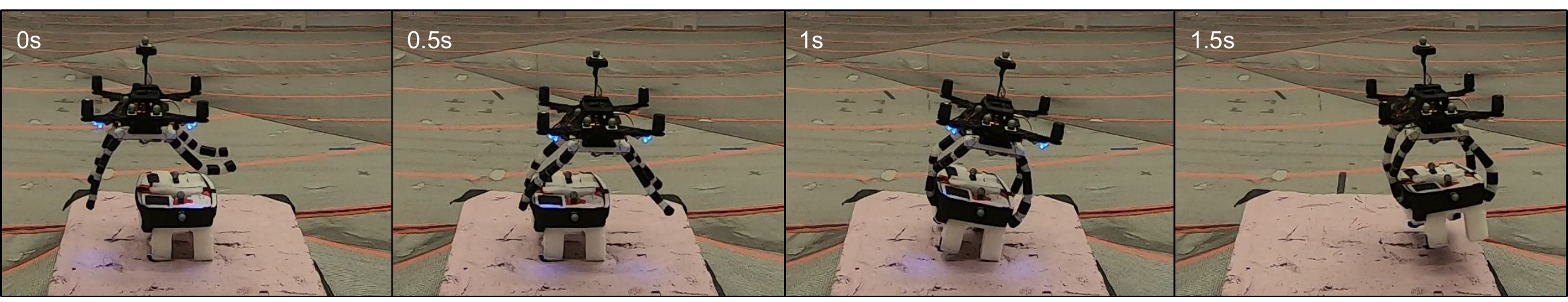}
    \caption{A closeup view of the grasping maneuver during a trial of the {dynamic grasping} tests. The gripper configuration for the ``approach'' phase mentioned in Section~\ref{sec:grasp_optimization_details} is visible at $t = \SI{0}{s}$. The drone visibly maintains consistent forward progress while grasping. The compliance of the fingers allows our \emph{\softDrone} to start contact with the grasp object before the gripper has enveloped the object fully; deformation of the back fingers is visible between the start of contact ($t = \SI{0.5}{s}$) and just before the gripper fully closes ($t = \SI{1.0}{s})$. The softness of the gripper also allows a more secure grasp by 
    conforming to the geometry of the target object ($t = \SI{1.5}{s})$.}\label{fig:grasp_real}
\end{figure*}


\myParagraph{Dynamic Grasping Results}
Fig.~\ref{fig:grasp_real} showcases our real system during dynamic grasping at $\SI{0.2}{m/s}$.
The physical restrictions of our first prototype --- including thrust saturation, state estimation delays, and soft gripper latency --- limit its maximum speed;
at the same time it enables us to provide a convincing example of dynamic grasping without the 
slow, precise positioning required by existing aerial manipulators. 
The tendon lengths computed by our trajectory optimization  
were $\{162, 190, 233, 208\}\,\SI{}{mm}$ for approach and $\{225, 225, 190, 190\}\,\SI{}{mm}$ for grasp (\cf Section~\ref{sec:aggressiveManipulation}).

\begin{table}[h!]
\centering
\begin{tabular}{ c | c | c | c }
    Total Trials & Successes  
    & Velocity at grasp \\
    \hline
    23 & 21 (91.3\%)  
    & \SI{0.2}{m/s} \\
\end{tabular}
    \caption{Soft drone performance in real {dynamic grasping}.}\label{tab:real_grasp_success}
\end{table}

Table~\ref{tab:real_grasp_success} reports statistics from 23 real tests (including those in Fig.~\ref{fig:controller_error_real}), while several examples are included in the video attachment.
The $91.3\%$ success rate confirms our soft drone can reliably perform dynamic grasping in the face of real-world disturbances. Moreover, our two observed failures were due to catastrophic state estimate divergence (causing the drone to crash instantly) and seem unrelated to grasping; we have observed similar errors in other tests without grasping and without the soft gripper attached. In this case they occurred at the end of the grasp trajectory on the fifth and tenth trial, causing the drone to crash while carrying the grasped object.

The 23 grasping trials took place consecutively across two days and were only interrupted to replace the batteries or reset the system after a crash.
Throughout, the \softDrone's performance was essentially unchanged. 
This consistency was despite adverse positioning errors which often resulted in at least one finger making no contact with the target (and the remaining fingers needing to adapt to compensate) -- exemplifying the advantages of \emph{morphological computing}.

We also attempted 5 grasping trials with the standard geometric controller, and none of them resulted in a grasp (\ie the drone passed too far  above the object). This is due to the fact that the geometric controller has larger tracking errors and is not able to compensate for the ground effect.  

\myParagraph{Limitations and Discussion} 
During real testing, we observed several phenomena that were not captured in the simulation of our system and that initially impacted our ability to grasp. 
Besides the ground effect that caused poor tracking performance for the drone in the vicinity of the object, we also observe a ``thrust stealing'' effect, where the grasped object would block airflow of the propellers. 

To address the ground effect, we also replicated the approach of~\cite{Shi19icra-neuralLander}, in which a compensatory force for the ground effect is learned from previous flights around the area of interest. We observed better tracking performance near the ground using this approach, but the learned model did not generalize to the reaction forces that the quadrotor encounters during grasping, leading to large errors after grasping.

Regarding the thrust stealing effect, we observed that 
keeping the mass of the target object the same while increasing its surface area resulted in the quadrotor not being able to stay in the air when moving between waypoints. We noticed also that more partial grasps led to worse trajectory tracking performance after grasping. 
 While the effects were exaggerated in our system due to how close we were to the maximum takeoff weight of the quadrotor, we believe 
 that the thrust stealing effect will deserve serious consideration in future systems. This is likely due to the fact that this effect 
---unlike the extra mass and moment the grasped object imparts on the system---
cannot be treated as a simple wrench disturbance that can be compensated for.


\section{Conclusion}
\label{sec:conclusion}

This paper presents the first prototype of a \emph{\softDrone}
--- a quadrotor where rigid landing gears are replaced with a 
soft tendon-actuated gripper to enable aggressive grasping.
We describe our \softDrone prototype, including design and fabrication, 
and review the set of algorithms we use for trajectory optimization and control. 
We evaluate the resulting system in photo-realistic physics simulations (based on SOFA and Unity) and in real tests in a motion-capture room.
Our drone is able to dynamically grasp objects of unknown shape where baseline approaches fail and with a success rate of $91.3\%$ across 23 trials.
 Our tests reveal 
 three elements that are key to successful dynamic grasping: 
 (i) the use of an adaptive controller (that can 
 compensate for the disturbances induced by the grasped object and unmodeled aerodynamic effects),
 (ii) the use of a trajectory optimization scheme to optimize the  
 gripper configuration during grasping, 
 (iii) the role of softness in mitigating the effect of tracking errors, dampening contact forces, and allowing the gripper to better conform to the target object during grasping.

While we demonstrate exciting initial results, 
pushing the envelop of aggressive grasping to meet the performance observed in nature will 
require further research on design, algorithms, and fabrication, and 
provides a formidable challenge for the robotics community.
We see two future challenges as potential catalysts for progress in this area:
 (i) aggressive grasping at speed higher than \SI{2}{m/s}, and (ii) aggressive grasping on 
 moving platforms without requiring the speed of the drone to match that of the moving platform, 
 a setup that has been described as ``unsafe to perform'' in conjunction to landing with a traditional quadrotor \cite{Alkowatly15jgcd}. 
\extended{
	\renewcommand{\baselinestretch}{0.90}
	{
		\tiny 
		\bibliographystyle{IEEEtran}
		\bibliography{./references/refs,./references/myRefs} %
	}
}
{
	\bibliographystyle{IEEEtran}
	\bibliography{./references/refs,./references/myRefs} %
}	

\end{document}